\documentclass[letterpaper, 10 pt, conference]{ieeeconf}  

\IEEEoverridecommandlockouts                              

\usepackage{float,graphicx}
\usepackage{caption}
\usepackage{trimclip}
\usepackage{amsfonts} 
\usepackage{amsmath}
\usepackage{amssymb}
\usepackage[margin=25mm]{geometry}
\usepackage[table,xcdraw]{xcolor}
\usepackage{booktabs}
\usepackage{multirow}
\usepackage{amsmath}
\usepackage{comment}
\usepackage[colorinlistoftodos]{todonotes}
\usepackage{cite}
\usepackage{float}
\usepackage{soul}
\usepackage{multirow}
\usepackage{bbding}
\usepackage{pifont}
\usepackage[nointegrals]{wasysym}
\usepackage{caption}
\usepackage{subcaption}
\usepackage{amsmath}
\usepackage{array}
\usepackage{color}
\usepackage[normalem]{ulem}
\usepackage{hyperref}
\usepackage{nomencl}
\makenomenclature
\usepackage[ruled,vlined]{algorithm2e}
\usepackage{amsmath,amssymb,amsfonts}
\usepackage{algorithmic}
\usepackage{graphicx}
\usepackage{textcomp}
\usepackage{wrapfig}
\usepackage{mdframed,lipsum}
\usepackage[T1]{fontenc} 
\usepackage{booktabs}
\usepackage{mwe}
\usepackage{wrapfig}

\title{\LARGE \bf
Imitation Learning based Auto-Correction of Extrinsic Parameters for A Mixed-Reality Setup 
}

\author{Shubham Sonawani, Yifan Zhou and Heni Ben Amor
}

\begin{document}
\makeatletter

\let\@oldmaketitle\@maketitle
\renewcommand{\@maketitle}{\@oldmaketitle
\centering
\includegraphics[width=\linewidth,height=8\baselineskip]
{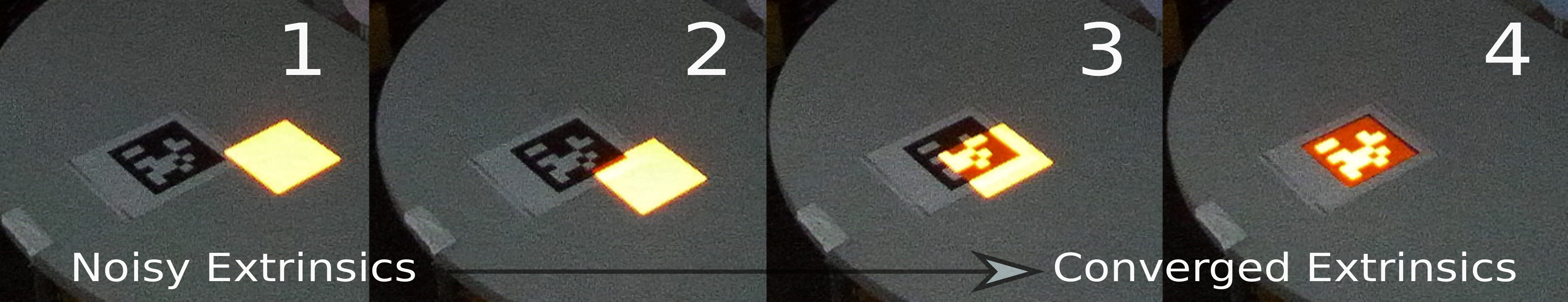}
\label{fig:intpro_arch}
}

\maketitle 
\thispagestyle{empty}
\pagestyle{empty}

\begin{abstract}
In this paper, we discuss an imitation learning based method for reducing the calibration error for a mixed reality system consisting of a vision sensor and a projector. Unlike a head mounted display, in this setup, augmented information is available to a human subject via the projection of a scene into the real world. Inherently, the camera and projector need to be calibrated as a stereo setup to project accurate information in 3D space. Previous calibration processes require multiple recording and parameter tuning steps to achieve the desired calibration, which is usually time consuming process. In order to avoid such tedious calibration, we train a CNN model to iteratively correct the extrinsic offset given a QR code and a projected pattern. We discuss the overall system setup, data collection for training, and results of the auto-correction model.  




\end{abstract}

\section{System Overview}
Mixed reality is receiving increased interest in robotics and human-robot interaction. In particular, it can be used to project visual signals to a human partner that can be easily and quickly understood. To this end, the work in \cite{ramsundar} proposed a camera-projector (stereo) mixed reality setup which provides explicit visual cues to a subject engaged in a human-robot collaborative task. Specifically, the approach used a pose information about an object of interest to generate and overlay visual information, e.g., highlighting an object, connecting two objects via a virtual line, etc. These signals were then projected into the real 3D scene to create an immersive environment combining real objects and artificial objects.

However, such systems require precisely calibrated extrinsic parameters in order to project visual cues perfectly onto real-world objects. The extrinsic identifies the transformation between the reference frame of camera and the projection device. Incorrect extrinsic parameters, as shown in an overview figure, can substantially affect the 3D effect and results in wrongly placed visual cues. In other words, it is not enough to identify the location of an object in the camera space alone, but rather its location in the projector frame is also critical. Given this, Deriving the extrinsic transformation between the camera and the projector is a non-trivial and delicate endeavor.

\begin{figure}[h!]
    \centering
    \includegraphics[scale=0.4]{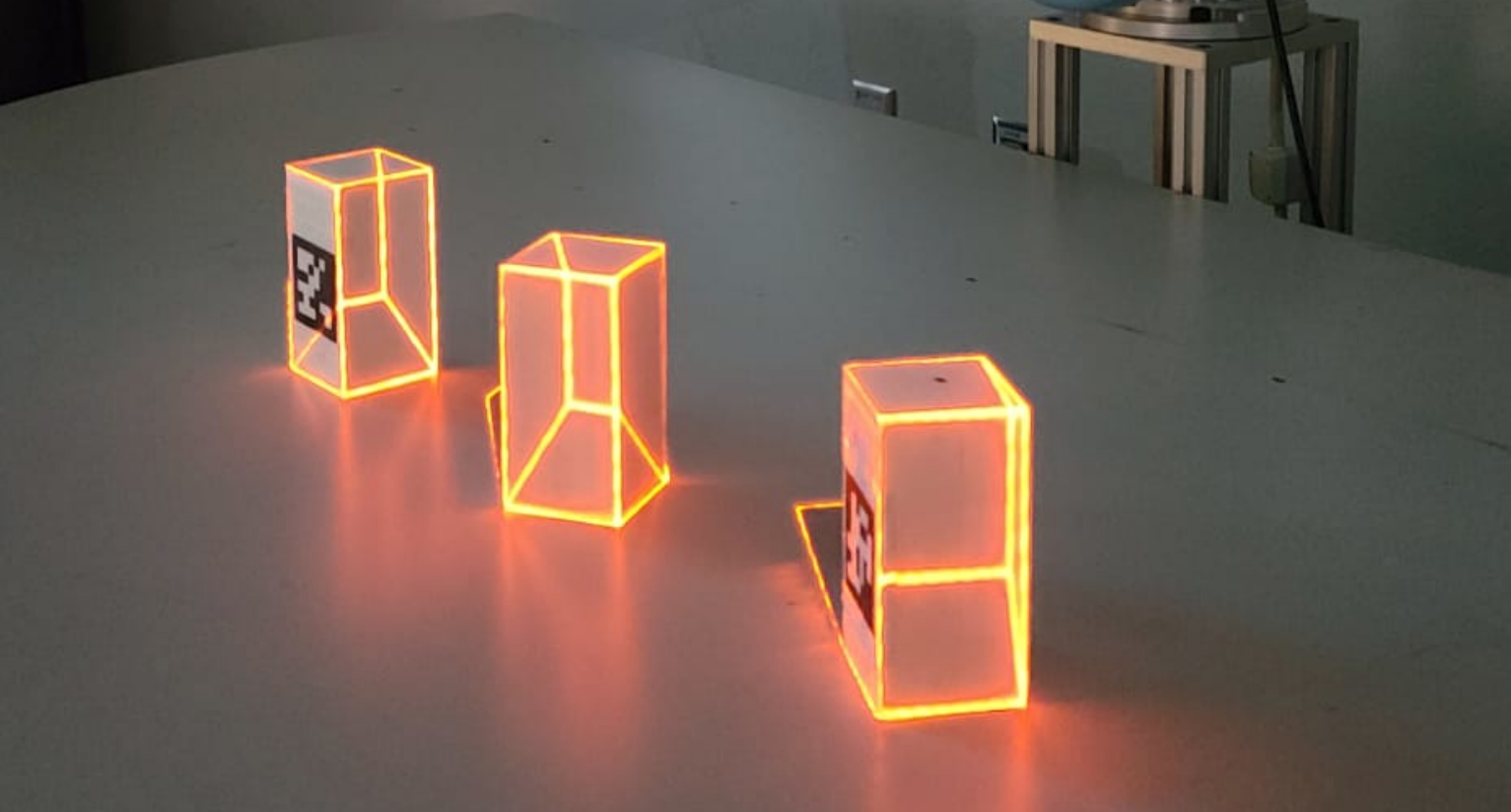}
    \caption{Projection of 3D cube's wire-frame after auto-calibrated extrinsic parameters}
    \label{fig:wireframe}
\end{figure}

For traditional camera calibration,  checkerboard-based calibration procedures are widely used and produce reasonably robust results. However, in the case of a project, the calibration procedure changes since a projector is an active light emitter and can be regarded as an inverse camera. As shown in \cite{Moreno2012}, the calibration sequence for a camera-projector setup requires tuning parameters such as global and projector-light components to generate a reasonable calibration. However, this adds an additional burden to the human expert performing the procedure and typically requires more than three image sequences to be recorded. Accordingly, the process is accurate but involves human intervention and can be time-consuming. Another element that is required is a physically large calibration pattern, i.e., a checkerboard that is big enough to capture a large section of the physical environment. Most importantly, any small change in the setup requires a complete recalibration -- a frustrating experience that can hamper the adoption of mixed reality.

To remove the requirement for such tedious processes, we propose a novel method that automatically uses a Neural Network model to automatically correct the extrinsic offset. In this method, we place an April tag (QR code) on the table and try to project a highlighted pattern onto the tag. Due to the inaccurate extrinsic parameters, the projected pattern will have an offset from the ground truth pose. By observing this offset visually, our model directly predicts the underlying extrinsic error to correct the extrinsic parameters. Gradually, the highlighted pattern shifts on top of the QR code. The overall result is that visual cues and patterns can now be accurately projected into space and on top of 3D objects to create complex and compelling mixed-reality environments, see Fig.~\ref{fig:wireframe}.

\section{Training and Testing of the Neural Network}
The goal of the model is to find a policy $\pi_{\pmb{\theta}}( \boldsymbol{e} | \boldsymbol{I})$ which is parameterized by $\boldsymbol {\theta}$, takes in an image $\boldsymbol{I}$ from the camera and estimates the current error $\boldsymbol{e}$ in the projector extrinsic parameters. In this image $\boldsymbol{I}$, there is a highlighted area and an April tag, where the highlight is supposed to cover the April tag perfectly. Since some error $\boldsymbol{e}$ exists in the extrinsic parameters, the highlighted area is offset from the intended projection area. Given this image, we train a neural network as the policy $\pi_{\pmb{\theta}}$ to directly predict the extrinsic parameters' offset. This is, in essence, an imitation learning process. The model is trained on demonstrations where experts adjust the offsets, and the model imitates these actions.

\subsection{The Neural Network Model}
We adopt a ResNet-50~\cite{He_2016_CVPR} neural network pre-trained on ImageNet-1k~\cite{5206848} dataset as our backbone. Since we are performing a regression task instead of a classification task as ResNet-50 originally does, we modify the last layer of the network to be a fully-connected layer that outputs the estimated offset of the extrinsic parameters directly. 

\subsection{Data Collection}
In order to train the network, we need a set of expert demonstrations $\mathcal{D} = \{\boldsymbol{d}_0, ..., \boldsymbol{d}_n\}$ where each demonstration $\boldsymbol{d}_i$ contains a current image $\boldsymbol{I}$ and the expert's estimation of correction $\boldsymbol{e}$. In order to collect the data, firstly, a camera projector system is calibrated tediously using \cite{Moreno2012} method with around eight recordings of calibration sequences. In addition, manual tuning of extrinsic parameters is required to obtain near perfect calibration parameters. 
We project 0.10 meters $\times$ 0.10 meters of a red colored plane onto the April tag tracked in the camera frame using \cite{olson2011apriltag}. In order to mimic the offset in extrinsic parameters, we inject random offset in the x and y position of the projector w.r.t. camera. Moreover, iteratively, in a sequence of images, we reduce the offset and capture an image at that offset showing the April tag and projected plane. We collect around 100 such sequences, of which we use 70 for training and 30 for testing.  

In order to enable high precision in the real world, we use high resolution images ($700 \times 700$ px), so that the images contain clear edges of the April tag and highlighted area.

\subsection{Testing in Real World}
For testing, we calibrate the system before starting the testing process to get the ground truth extrinsic parameters for comparison. Afterward, we initialize a test trial by randomly placing the April tag on the table. While testing, we disturb the ground truth extrinsic parameters and use them as the noisy extrinsic parameters, which causes an offset on the highlighted pattern (see overview figure with the convergence of noisy to ground truth sequence 1-4). The trained neural network takes in the image captured by the camera and spits out the estimated offset of the extrinsic parameters, with which the correctional update is made, and the pattern is projected back onto the table with lower error. Repeating this step iteratively, we can see the highlighted pattern is moving closer to the April tag (overview figure parts 2 and 3). When the predicted offsets are smaller than a pre-defined small number ($\epsilon$), we mark the convergence of the trial. Empirically, we choose a small step size for updates, which smoothes out the potential prediction noises and increases the stability of the convergence.

We measure the l-2 norm between the ground truth and the converged prediction as the error. During test time, the error is around 5e-4. Corresponding to the real world set up, we can see the highlight pattern is projected onto the April tag precisely (checkout the overview figure-part 4).

\section{Conclusion}
This paper demonstrates work in progress auto-calibration method for a camera-projector based mixed reality setup. Importantly, with 100 data sequences for training and testing, we show that the CNN model can be used to calibrate extrinsic parameters of the camera-projection stereo system. However, this paper has not addressed convergence guarantees of CNN model to ground truth parameters. Given this, future work will involve analysis of convergence guarantees and robustness of the discussed method. 

\bibliographystyle{ieeetr}
\bibliography{references}

\end{document}